%% file: main.tex
\documentclass[lettersize,journal]{IEEEtran}
\usepackage[utf8]{inputenc}
\usepackage{amsmath,amsfonts}
\usepackage{algorithm}
\usepackage{algpseudocode}
\usepackage{array}
\usepackage[caption=false,font=normalsize,labelfont=sf,textfont=sf]{subfig}
\usepackage{textcomp}
\usepackage{stfloats}
\usepackage{url}
\usepackage{verbatim}
\usepackage{graphicx}
\usepackage{cite}
\usepackage{ulem}
\usepackage{comment}
\usepackage[ngerman,english]{babel}
\usepackage{paralist, tabularx}
\usepackage[dvipsnames]{xcolor}
\usepackage{multirow}
 \usepackage[utf8]{inputenc}  
\usepackage{booktabs}
\usepackage{hyperref}

\usepackage{graphicx}
\usepackage{float}
\usepackage{booktabs}
\usepackage{pifont}
\usepackage{algpseudocode}
\usepackage{amsmath}
\usepackage{amssymb}
\usepackage{mathtools}
\usepackage{amsthm}
\usepackage{enumitem}
\usepackage{bm}
\usepackage[capitalize,noabbrev]{cleveref}
\usepackage{colortbl}
\usepackage{tcolorbox}
\definecolor{light-gray}{gray}{0.94}
\usepackage{xcolor} 
\usepackage{caption}
\newif\ifmodify 
\ifmodify

\else

\fi

\begin{document}
\twocolumn
\title{
TSLA: A Task-Specific Learning Adaptation for Semantic Segmentation on Autonomous Vehicles Platform
}

\author{
Jun Liu, Member, IEEE,
Zhenglun Kong$^\dagger$,
Pu Zhao,
Weihao Zeng,
Hao Tang,
Xuan Shen, 
Changdi Yang,\\
Wenbin Zhang,
Geng Yuan, Member, IEEE,
Wei Niu,
Xue Lin
and 
Yanzhi Wang, Member, IEEE

\IEEEcompsocitemizethanks{\IEEEcompsocthanksitem J. Liu, Z. Kong, P. Zhao, X.Shen, C.Yang, X. Lin, and Y. Wang are with the Department
of Electrical and Computer Engineering, Northeastern University, Boston, MA, 02115. W. Zeng, H. Tang are with the Robotics Institute, Carnegie Mellon University,  Pittsburgh, PA, 15213. G. Yuan, W.Niu are with the School of Computing at the University of Georgia, Athens, GA 30602. W.Zhang is with the Knight Foundation School of Computing \& Information Sciences at Florida International University, Miami, FL 33199.\protect\\ \vspace{-0.23in}

}
\thanks{$\dagger$ Corresponding author: Zhenglun Kong, kong.zhe@northeastern.edu}

}



\maketitle

\input{sections/0-abstract}

\input{sections/1-keywords}

\input{sections/2-introduction}

\input{sections/3-related-work}

\input{sections/5-method}

\input{sections/6-exp-results}

\input{sections/7-conclusion}

\bibliographystyle{IEEEtran}
\bibliography{IEEEabrv, reference}

\end{document}

%% file: sections/0-abstract.tex
\begin{abstract}

Autonomous driving platforms encounter diverse driving scenarios, each with varying hardware resources and precision requirements. Given the computational limitations of embedded devices, it is crucial to consider computing costs when deploying on target platforms like the DRIVE PX 2. Our objective is to customize the semantic segmentation network according to the computing power and specific scenarios of autonomous driving hardware. We implement dynamic adaptability through a three-tier control mechanism—width multiplier, classifier depth, and classifier kernel—allowing fine-grained control over model components based on hardware constraints and task requirements. This adaptability facilitates broad model scaling, targeted refinement of the final layers, and scenario-specific optimization of kernel sizes, leading to improved resource allocation and performance.

Additionally, we leverage Bayesian Optimization with surrogate modeling to efficiently explore hyperparameter spaces under tight computational budgets. Our approach addresses scenario-specific and task-specific requirements through automatic parameter search, accommodating the unique computational complexity and accuracy needs of autonomous driving. It scales its Multiply-Accumulate Operations (MACs) for Task-Specific Learning Adaptation (TSLA), resulting in alternative configurations tailored to diverse self-driving tasks. These TSLA customizations maximize computational capacity and model accuracy, optimizing hardware utilization.

\end{abstract}

%% file: sections/1-keywords.tex
\def\IEEEkeywordsname{Keywords}

\begin{IEEEkeywords}

Scenario-Specific-Task-Specific; auto adjustable convolutional kernels; scalable depth multiplier; classifier depth; kernel depth;  flexible computational complexity; MobileNetV4

\end{IEEEkeywords}

%% file: sections/2-introduction.tex
\section{Introduction}

Real-time scene understanding is essential for perception in mobile robotics and self-driving cars. Semantic segmentation, which classifies each pixel in an image into categories like 'road' or 'sky,' is crucial for assisting in localization and planning for informed decision-making.

\begin{figure}[t]
  \centering
  \includegraphics[width=0.92\linewidth]{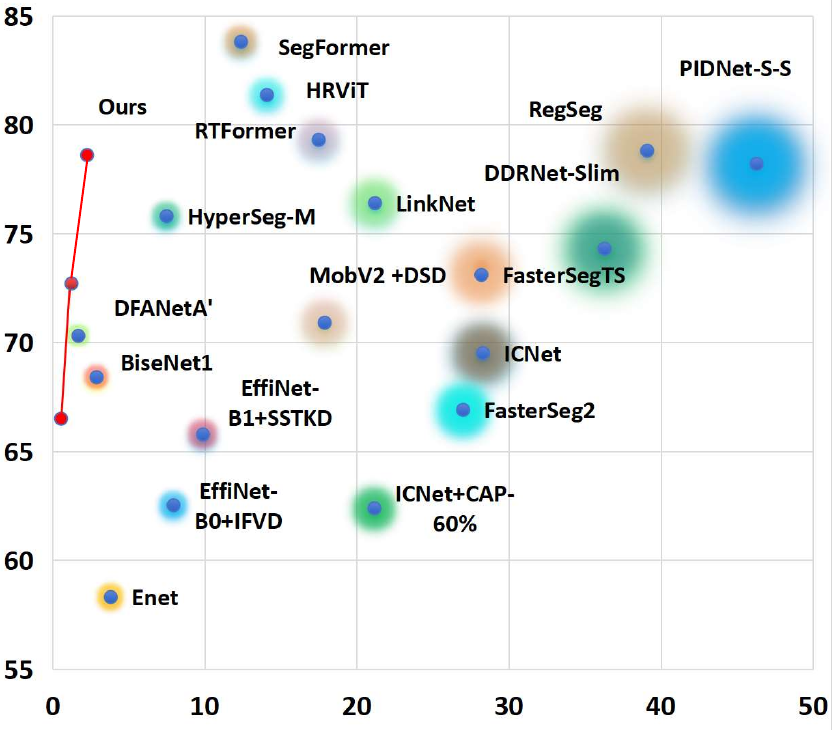}
  \caption{
 Comparison of FLOPs and mIoU on cityscapes test set with real-time methods. The bigger the bubble, the larger the computational complexity of the model. The horizontal axis is mIoU, and the vertical axis is GFLOPS.
  }
  \label{fig:results_compare}
\vspace{-5mm}
\end{figure}

Semantic segmentation with convolutional neural networks (CNNs) can be computationally expensive, especially with large datasets or complex architectures. Training and deploying CNNs often require significant computational resources, making it a major challenge. Previous research has explored various strategies to address these computational demands.
Pruning methods~\cite{li2025mutual,yuan2021work,liu2025toward,zhao2024pruningfoundationmodelshigh,li2022pruning,zhang2022advancing} may require careful fine-tuning to balance model size reduction and performance degradation. 
Run-time pruning introduces additional complexity and overhead during inference, which may impact real-time performance. 
Quantization techniques~\cite{han2015deep} struggle to meet the diverse requirements of autonomous driving scenarios, due to the fixed bit-width of weights in quantized models.
Neural Architecture Search (NAS) ~\cite{10.1145/3528578,shen2024search,wu2022compiler,zhan2021achieving} techniques can be computationally expensive themselves, requiring extensive search and evaluation. 
Knowledge distillation~\cite{liu2025structured,li2019dfanet} often requires access to a pre-trained complex model, limiting its applicability in certain scenarios. 
Vision Transformer (ViT)~\cite{dosovitskiy2020image,liu2025rora} has shown impressive accuracy in semantic segmentation tasks. However, it is important to note that ViT models often demand significant computing power due to their complex architecture. For instance,  SegFormer~\cite{xie2021segformer}, achieves state-of-the-art mean Intersection over Union (mIoU) in semantic segmentation. However, this achievement comes with the drawback of high computational requirements, demanding 8.4 billion floating-point operations (flops) for processing. 


Autonomous driving scenarios cover a broad spectrum of driving situations, each characterized by distinct road conditions, traffic scenarios, and environmental factors. These scenarios require specific tasks and performance criteria. To achieve optimal performance, autonomous driving manufacturers often employ different recognition models that are specifically designed for different scenarios. This approach ensures that the semantic segmentation models are fine-tuned and optimized to perform effectively in each unique driving situation. 
To fulfill the requirements, our aim is to customize the semantic segmentation network, specifically designed for Scenario-Specific-Task-Specific scenarios, to align with the computing power of the autonomous driving hardware.
To achieve this objective, we make modifications to the best network architectures suitable for edge devices, specifically MobileNetV4~\cite{qin2024mobilenetv4}. These adaptations enable us to optimize the semantic segmentation network for deployment on autonomous driving hardware, aligning it with the computing power and efficiency requirements of edge devices. Evaluation of their proposed models is performed using the CamVid dataset~\cite{BrostowSFC:ECCV08} and Cityscapes~\cite{7298965}. By fine-tuning the network's computational demands and capacity, a harmonious equilibrium can be achieved between precision and effectiveness, aligning with the particular demands of our use case on the embedded platform. 

Overall, this article introduces the following contributions:
\begin{itemize}[leftmargin=*]
   \item  We are the first to leverage MobileNetV4 for constructing a semantic segmentation network tailored for edge devices. We implement dynamic adaptability through a three-tier control mechanism—width multiplier, classifier depth, and classifier kernel—allowing fine-grained control over model components based on hardware constraints and task requirements. This adaptability facilitates broad model scaling, targeted refinement of the final layers, and scenario-specific optimization of kernel sizes, leading to improved resource allocation and performance.
   \item  We leverage Bayesian Optimization with surrogate modeling to efficiently explore hyperparameter spaces under tight computational budgets. Our approach addresses scenario-specific and task-specific requirements through automatic parameter search, accommodating the unique computational complexity and accuracy needs of autonomous driving. It scales its Multiply-Accumulate Operations (MACs) for Learning Adaptation (TSLA).
   \item Our framework is designed for real-world scenarios like autonomous driving and embedded systems, requiring models to adapt in real-time to hardware limitations and evolving task demands. 
\end{itemize}  

In general, as illustrated in Fig.~\ref{fig:results_compare}, these contributions enable customization of the MobileNet architecture to align with distinct deployment prerequisites and enhance efficiency while accounting for computational constraints. Our approach offers the capability to modify the model's computational intricacy and capability, ranging from 0.57 to 1.90 GFLOPS, catering to diverse autonomous driving scenarios.

%% file: sections/3-related-work.tex
\section{Related Work}

Modern Convolutional Neural Networks (CNNs) require substantial computational resources, often involving millions to billions of operations per prediction and even higher during training. By using equations to estimate the computational load of different layers, we can accurately assess the demands of common CNN architectures, offering a key metric for hardware performance evaluation.

Performance in semantic segmentation is often measured by peak accuracy during real-time testing on the Cityscapes dataset. Fig.~\ref{fig:results_compare} shows that while superior network architectures achieve higher mIOU, they also demand more computational resources, as indicated by Giga operations on the horizontal axis.
\textcolor{black}{In the visual representation, each data point is depicted as a distinct bubble, with the size of the bubble indicating the model's MACs (Multiply-Accumulate Operations. A greater bubble size means more computing power is required.} 
This graphical representation aids in selecting an optimal network structure by balancing computational efficiency and precision for specific tasks. It highlights that widely used architectures like  GoogLeNet~\cite{szegedy2015going} may not perform optimally under limited computational constraints.

\subsection{Efficient Lightweight Methods}
Researchers continually advance semantic image segmentation models for real-time applications through innovative methods and architectures~\cite{liu2023scalable,liu2023interpretable,liu2022efficient}.
PIDNet~\cite{sun2020pidnet} features three branches for detailed, contextual, and boundary data analysis, incorporating boundary attention to merge these branches effectively. ShuffleNet~\cite{zhang2018shufflenet} promotes cross-channel interactions through channel shuffling, enhancing feature diversity. DDRNet~\cite{hong2021deep} uses dual branches with bilateral fusions, while Enet~\cite{paszke2016enet} performs early downsampling to limit feature maps. ICNET~\cite{zhao2018icnet} employs a multi-resolution approach with cascade feature fusion for faster high-quality segmentation. HyperSeg~\cite{nirkin2021hyperseg} uses a nested U-Net for higher-level contextual features and dynamic patch-wise convolutions. SFNet~\cite{li2020semantic} introduces a Flow Alignment Module for improved semantic flow between feature maps. RTFormer~\cite{wang2022rtformer} leverages a transformer with dual-resolution efficiency for real-time segmentation. MobileNetV4~\cite{qin2024mobilenetv4} features the Universal Inverted Bottleneck (UIB) search block, unifying Inverted Bottleneck (IB), ConvNext, Feed Forward Network (FFN), and a new Extra Depthwise (ExtraDW) variant.

\begin{table*}[tb!]
    \centering
    \caption{The target hardware for our project is the DRIVE PX 2 platform. Our main objective is to fulfill a specific set of requirements for one of the three tasks mentioned previously. } 
    \resizebox{\linewidth}{!}{
    \begin{tabular}{l|c|c|c|c|c}
     \toprule
        \textbf{Scenario} &\textbf{Number of classes} & \textbf{Number of cameras} &  \textbf{Frame rate for processing} &  \textbf{Required accuracy} &  \textbf{Computational budget}  \\
        \midrule
         \multirow{2}{*}{Parking }   &7 classes: Sky, Infrastructure, Road, Pavement, &\multirow{2}{*}{4 cameras} &\multirow{2}{*}{Medium frame rate: 15 fps} &\multirow{2}{*}{Medium accuracy}  &Low budget on platform 70 Giga \\
        &Car, Vulnerable Road Users, Unlabeled, & & &  & operations per second\\
        \midrule
         \multirow{2}{*}{Urban}   &7 classes: Sky, Infrastructure, Road, Pavement, &\multirow{2}{*}{4 cameras} &\multirow{2}{*}{High frame rate: 30 fps} &\multirow{2}{*}{High accuracy}  &High budget on platform 300 Giga \\
        &Car, Vulnerable Road Users, Unlabeled, & & &  & operations per second\\
        \midrule
        \multirow{2}{*}{\parbox{1.0cm}{Rural }} &\multirow{2}{*}{2 classes: Road, Not Road} &\multirow{2}{*}{1 camera} &\multirow{2}{*}{High frame rate: 30 fps} &\multirow{2}{*}{High accuracy} &Medium budget on platform 120 Giga\\
       & & && & operations per second\\
    \bottomrule
    \end{tabular}
   }
   
   \label{tbl:three_basic}
\vspace{-3mm}
\end{table*}

\subsection{Pruning}
Pruning~\cite{10.1145/3528578,rtseg,zheng2024exploring,yang2023pruning,shen2024edgeqat} reduces the complexity of deep learning models by eliminating certain parameters or connections to enhance efficiency and lower computational and memory requirements while preserving performance. Various techniques use criteria such as weight magnitude or connection importance to guide the pruning process~\cite{li2020efficient}. Pruning can occur during training or as a post-processing step, focusing on removing weights with minimal impact or connections that have little effect on the model output. 

Run-time pruning~\cite{li2022bitxpro,kong2023peeling} dynamically reduces computational load and memory during inference by adjusting the network structure based on input data. Unlike traditional training-time pruning, it removes less important neurons, channels, or filters in real-time, making it ideal for deploying lightweight models in resource-constrained environments.

\subsection{Knowledge Distillation}
Knowledge distillation~\cite{li2019dfanet} is a deep learning strategy for model compression, where a compact student model is trained to replicate the performance of a larger teacher model. This enables the student model to achieve similar results with better computational efficiency. 
In knowledge distillation, the student model is trained using the teacher model's soft probabilities (logits) rather than one-hot labels. These soft probabilities indicate the teacher's confidence in each class and are especially useful for resource-constrained devices or where computational efficiency is critical.
\subsection{Quantization}
Quantization~\cite{han2015deep} in deep learning involves reducing the precision of model parameters and activations by representing numbers with fewer bits, shifting from floating-point (32-bit or 16-bit) to fixed-point formats (8-bit, 4-bit, or lower)\cite{hubara2018quantized}. Techniques include Weight Quantization, which reduces weight precision while keeping activations higher, Activation Quantization, which reduces activation precision while keeping weights higher, and Full Quantization, which quantizes both weights and activations. While quantization helps save resources, it can lead to accuracy loss due to reduced precision. Thus, the challenge is to balance precision reduction with maintaining model performance. 

\subsection{Neural Architecture Search (NAS)}
Neural Architecture Search (NAS)~\cite{10.1145/3528578} automates the design of optimal neural network architectures by exploring a predefined search space and selecting the best-performing models based on specific metrics~\cite{dong2023speeddetr,dong2024hotbev}. It employs search algorithms like genetic algorithms, reinforcement learning, Bayesian optimization, and gradient-based methods to navigate and evaluate different architectures. 

%% file: sections/5-method.tex
\section{METHOD}

\begin{figure*}[t]
  \centering
  \includegraphics[width=0.9\linewidth]{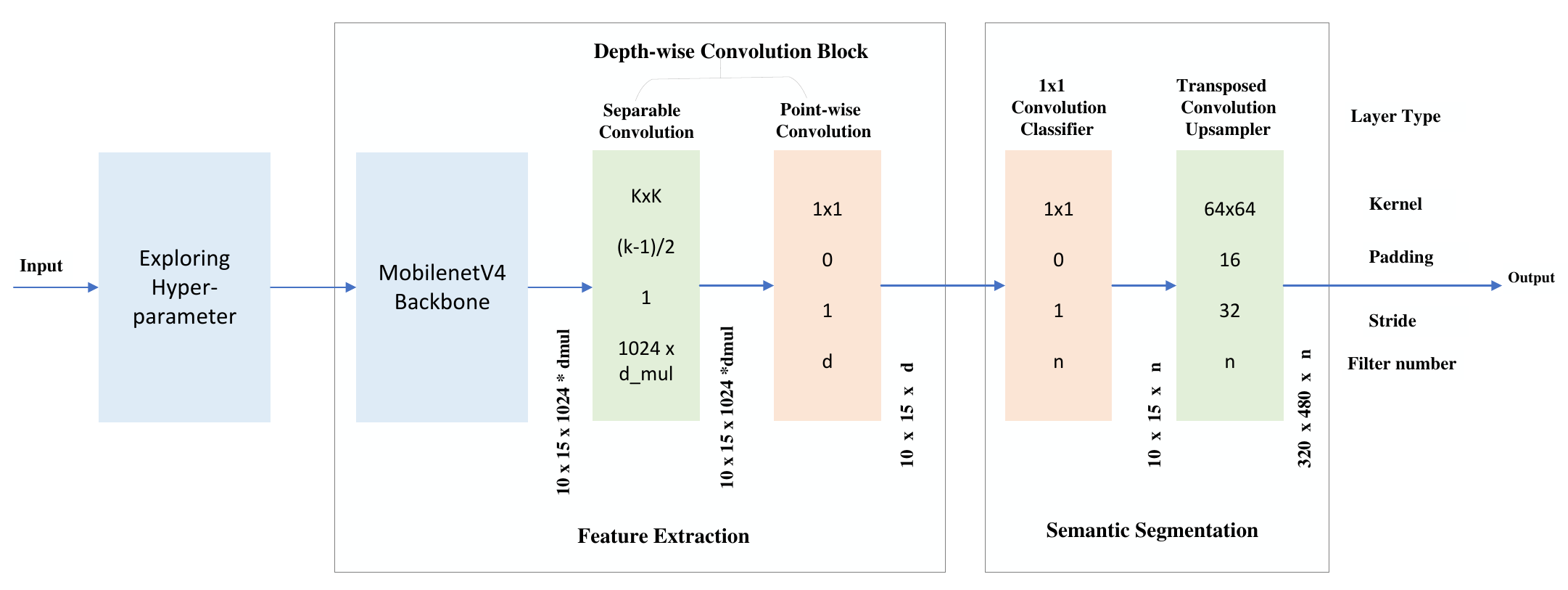}
  \caption{
 The architecture of the Task-Specific Learning Adaptation (TLSA) network consists of two main blocks. The left block serves as the Feature Extractor, with a kernel size denoted as $k$ and a width multiplier~\cite{DBLP:journals/corr/HowardZCKWWAA17} denoted as $d\_mul$. The right block is responsible for Semantic Segmentation, with the number of classes denoted as $n$. In addition, the classifier depth is represented as $d$.
  }
  \label{fig:tsla}
\vspace{-4mm}
\end{figure*}

\subsection{Scenarios}  

Autonomous driving involves diverse scenarios, each with unique road conditions, traffic, and environmental challenges. To meet the specific demands of semantic segmentation, models must be adaptable to different tasks, ensuring real-time, accurate object recognition and scene analysis. This adaptability allows models to handle the dynamic nature of driving environments, responding effectively to varying conditions.

By adapting to specific scenarios, semantic segmentation models can enhance the capabilities of autonomous vehicles, enabling safe navigation, informed decision-making, and effective environmental perception. As shown in Table~\ref{tbl:three_basic}, scenario selection and parameter justification should consider the following factors:

\noindent{\textbf{Scenario Selection.}}

\begin{itemize}[leftmargin=*]
  \item Parking: 
This scenario typically involves low-speed maneuvers in confined spaces, where the vehicle needs to detect various objects such as infrastructure, other cars, and pedestrians. The need for medium frame rates and lower computational budgets reflects the lower dynamic nature of this scenario.

  \item Urban:
Characterized by high traffic density, multiple types of road users, and complex interactions. High frame rates and high accuracy are crucial to handle the fast-paced and unpredictable nature of urban driving.

  \item Rural:
Rural roads are generally less complex than urban areas, but require high precision to distinguish between road and non-road regions due to the higher speeds involved. The computational budget is set medium to balance between accuracy and efficiency.
\end{itemize}
\noindent{\textbf{Parameter Justification.}}
 \begin{itemize}[leftmargin=*]
  \item Number of Cameras: 
The number of cameras is determined by the complexity of the environment and the necessity to cover blind spots. Urban and Parking scenarios require more cameras to capture the environment fully, while a single camera suffices for the simpler Rural scenario.

  \item Frame Rate:
Frame rate requirements are directly related to the speed and complexity of the environment. High frame rates are essential in fast-moving scenarios (Urban and Rural) to ensure timely processing of information.

  \item Computational Budget:
Designed to match the processing demands of the scenario. Urban scenarios demand more processing power due to the higher number of detected classes and the need for high accuracy.
\end{itemize}

\color{black}
In Table~\ref{tbl:three_basic}, for autonomous driving manufacturers, models need to meet the requirements of at least three basic scenarios for autonomous driving.

\color{black}

\subsection{Model Design}  
MobileNet offers a more flexible architecture that is particularly suited for balancing accuracy and computational cost, which is crucial in resource-constrained environments like embedded systems for autonomous driving. Its adjustable width multiplier and classifier depth make it highly customizable for various hardware and task-specific requirements.

While the point-wise convolution in MobileNet's depthwise separable convolution does not involve cross-channel information interaction, this is addressed in MobileNetV4 through the introduction of the Universal Inverted Bottleneck (UIB). The UIB not only enables more efficient channel mixing but also extends the receptive field and improves computational efficiency by focusing on hardware-friendly operations~\cite{qin2024mobilenetv4}.

ShuffleNet uses a channel shuffle operation to enable efficient information exchange between grouped convolutions, enhancing channel interaction. However, this method introduces additional computational complexity~\cite{zhang2018shufflenet}. While effective, ShuffleNet lacks the same flexibility in architectural customization as MobileNet, especially for applications requiring dynamic adjustments to meet diverse hardware and task requirements.

For our objectives-optimizing both accuracy and efficiency on autonomous driving platforms like DRIVE PX2-MobileNetV4, with its simpler design and adaptability to computational constraints, is a better fit for Table~\ref{tbl:three_basic}. While ShuffleNet is powerful, it introduces unnecessary complexity for our use case. 

Previous approaches like standard MobileNet and similar architectures focus on uniform model scaling, lacking dynamic fine-tuning for hardware constraints or task-specific needs. Our approach provides three-tier adaptability fine-grained control through:

\begin{itemize}[leftmargin=*]
    \item \underline{Broad model scaling} via WIDTH MULTIPLIER, offering flexibility in adjusting network depth, absent in fixed-size models.
    \item \underline{Targeted refinement} using CLASSIFIER DEPTH for specific tuning of the final layers, crucial for tasks like autonomous driving.
    \item \underline{Scenario-specific optimization} through CLASSIFIER KERNEL, adjusting kernel sizes for tasks in real-time edge computing.
\end{itemize}

Unlike MobileNet, which typically uses \textbf{fixed kernel sizes} in depthwise convolutions, our approach introduces \textbf{dynamic kernel size adjustments}, providing more control over feature extraction. This flexibility is critical for \textbf{autonomous driving platforms}, where varying driving scenes demand different kernel sizes to effectively capture specific features, optimizing both computational resources and performance.
Here are the key advantages of incorporating adjustable kernel sizes:
\begin{itemize}[leftmargin=*]
    \item \underline{Enhanced Feature Extraction:} Adjustable filters improve feature extraction by optimizing for specific input features and capturing various scales and complexities.
    \item \underline{Flexibility and Adaptability:} The ability to vary kernel sizes improves the network's flexibility, allowing it to handle different types of input data in dynamic environments, such as autonomous driving scenarios.
    \item \underline{Reduced Computation and Parameters:} While separable convolutions already reduce computation and the number of parameters, adjustable filters further minimize computational costs and memory usage while maintaining or enhancing performance, which is critical for embedded devices.
\end{itemize}
In Figure~\ref{fig:tsla},  Our TLSA architecture consists of a feature extractor (built on a MobileNetV4 backbone) and a semantic segmentation module. Our Separable Convolution Layer introduces the ability to adjust kernel sizes (k), overcoming MobileNet’s limitations in this regard. This adjustability is a key feature of our approach and enables us to balance computational efficiency with performance, particularly on resource-constrained hardware such as autonomous driving platforms. Thus, while MobileNet’s depthwise separable convolutions provide efficiency, our method extends this by enabling more \textbf{granular control} over filter sizes, enhancing its adaptability for task- and scene-specific requirements. 
\color{black}
The Semantic Segmenter consists of a classifier implemented by a 1x1 Convolution Layer and an Upsampler implemented by a Transposed Convolution Layer. 


\subsection{Adjustable Feature Extractor}  \label{sec:feature}
\noindent{\textbf{Depth-wise Convolutions Operation.}} With MobileNets, which is a collection of networks rather than a single deep neural network (DNN), a flexible and scalable approach for designing convolutional networks at various levels of complexity is employed. This allows the network architecture to be tailored to meet the specific demands of the task. Depending on the requirements of the task at hand, A MobileNet variant that strikes the right balance between computational efficiency and accuracy, making it suitable for a wide range of applications is chosen.

\begin{figure}[t]
  \centering
  \includegraphics[width=1.0\linewidth]{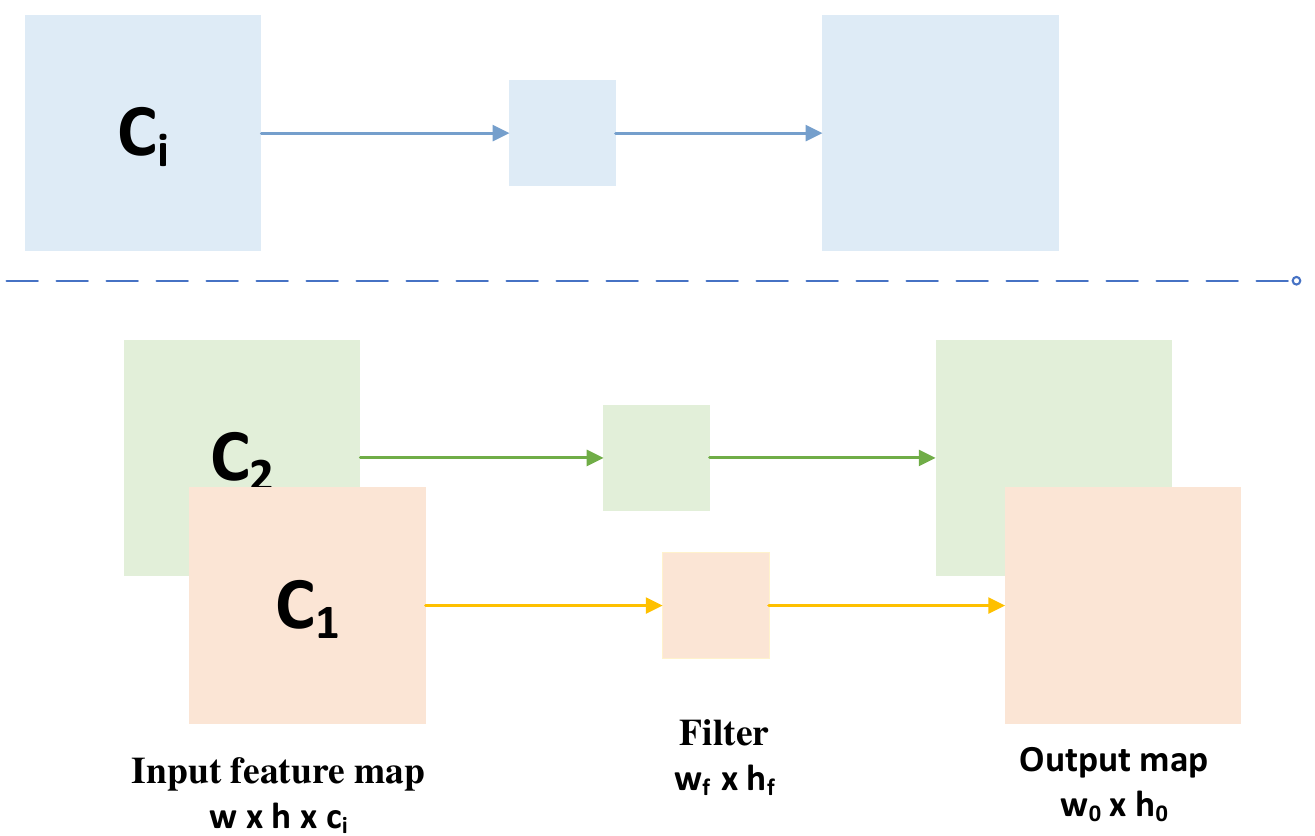}
  \caption{
Depth-wise convolutions.
  }
  \label{fig:depth-wise}
\vspace{-4mm}
\end{figure}

As depicted in Figure~\ref{fig:depth-wise}, depth-wise convolution uses a 3 × 3 kernel to convolve each input feature map independently, without processing across the full depth of all input feature maps. This step, known as Separable Convolutions (SC), produces the same number of output feature maps as the input, without mixing them. The computation involved in Depth-wise Convolutions can be quantified using the term: 
\begin{equation}
\text{MAC}^{sc} = h_o \times w_o \times c_i \times h_f \times w_f,
\end{equation}
where the $\text{MAC}^{sc}$ represents the number of multiply-accumulate operations for separable convolution, while $h_{\text{o}}$ and $w_{\text{o}}$ represent the output height and width, $c_{\text{i}}$ represents the number of input channels, and $h_{\text{f}}$ and $w_{\text{f}}$ represent the filter height and width, respectively.

To address the limitations of Separable Convolutions, which restrict the selection of output feature maps and channel mixing, Point-wise Convolutions are introduced. These 1×1 convolutions follow the Separable Convolution, operating across the output feature maps. The same formula for calculating MACs in Classical Convolution is applied here, with kernel height and width set to 1, ensuring that the output resolution remains $h_{\text{o}}$ x $w_{\text{o}}$. $c_{\text{i}}$ is maintained from the previous $\text{MAC}^{sc}$ calculation, and $n_{\text{f}}$ represents the number of output feature maps for Point-wise Convolution. The formula for calculating MACs for Point-wise (PW) Convolution is as follows:
\begin{equation}
\text{MAC}^{pw} = h_{o} \times w_{o} \times c_{i} \times n_{f} \times 1 \times 1 = h_{o} \times w_{o} \times c_{i} \times n_{f},
\end{equation}
where $h_{\text{o}}$ represents the output height, $w_{\text{o}}$ represents the output width, $c_{\text{i}}$ represents the number of input channels, and $n_{\text{f}}$ represents the number of filters. The multiplication by "1" in the equation indicates that the MAC operation involves multiplying by a scalar value of 1, which does not affect the overall result.

Therefore, Depthwise Separable Convolution require the following number of MACs and operations:
\begin{equation}
\begin{aligned}
\text{MAC}^{\text{DSP}} &= \text{MAC}^{\text{SC}} + \text{MAC}^{\text{PW}} \\
&= h_{o} \times w_{o} \times c_{i} \times h_{f} \times w_{f} + h_{o} \times w_{o} \times c_{i} \times n_{f} \\
&= h_{o} \times w_{o} \times c_{i} \times (h_{f} \times w_{f} + n_{f}),
\end{aligned}
\label{eq:macs}
\end{equation}

\begin{equation}
\begin{aligned}
\text{ops}^{\text{DSP}} &= 2 \times \text{MAC}^{\text{DW}}\\
&= 2 \times h_{o} \times w_{o} \times c_{i} \times (h_{f} \times w_{f} + n_{f}).
\end{aligned}
\label{eq:ops}
\end{equation}

The reduction factor (rops) of operations for using Depth-wise Convolutions instead of Classical Convolution is:

\begin{equation}
\begin{aligned}
\text{rops} &= \frac{2 \times h_{o} \times w_{o} \times c_{i} \times h_{f} \times w_{f} \times n_{f}}{2 \times h_{o} \times w_{o} \times c_{i} \times (h_{f} \times w_{f} + n_{f})} \\
&= \frac{h_{f} \times w_{f} \times n_{f}}{h_{f} \times w_{f} + n_{f}}.
\end{aligned}
\end{equation} 

This reduction factor indicates how much more efficient Depth-wise Convolutions are compared to Classical Convolutions, and the determination is based on the size of the kernel ($h_{\text{f}}$ x $w_{\text{f}}$) and the quantity of output feature maps $n_{\text{f}}$).

\noindent{\textbf{Width Multiplier in MobileNets.}} The Width Multiplier \color{black}is a parameter used in MobileNet instances to control the number of filters per Depth-wise Convolution layer. By adjusting the Width Multiplier, the number of filters can be decreased or increased based on our specific requirements. The default depth is set to 1024, and the way the number of MACs scales with different Width Multipliers is observed.

\begin{table}[t!]
    \vspace{0mm}
    \centering
    \caption{
    The number of multiply-accumulates scales with different width multipliers.} 
    \label{tbl:macs}
    \resizebox{\linewidth}{!}{
    \begin{tabular}{c|c}    
     \toprule
        \textbf{Width} &\textbf{MACs count of the example Depth-wise}  \\
        \textbf{Multiplier} &\textbf{ Convolution layer}  \\
        \midrule
        8/1024    &{4 × 4 × 5 × (3 × 3 + 8) = 1,360}\\
        0.25  &{4 × 4 × 5 × (3 × 3 + 256) = 21,200}      \\ 
        0.5  &{4 × 4 × 5 × (3 × 3 + 512) = 41,680} \\
        0.75  &{4 × 4 × 5 × (3 × 3 + 768) = 62,160} \\
        1.0  &{4 × 4 × 5 × (3 × 3 + 1024) = 82,640} \\
        1.25  &{4 × 4 × 5 × (3 × 3 + 1280) = 103,120} \\
        2.0  &4 × 4 × 5 × (3 × 3 + 2048) = 164,560 \\
    \bottomrule
    \end{tabular}
   }
\vspace{-3mm}
\end{table}

Table~\ref{tbl:macs} illustrates the variation in MACs count with different Width Multipliers, as determined by formula (5). The values used in the calculation are as follows:
\begin{itemize}
\item Height = 4 (the height of the feature map)
\item Width = 4 (the width of the feature map)
\item Channels = 5 (the number of input channels)
\item Width Multiplier \color{black} = 0.25, 0.5, 0.75, 1.0, 1.25, 2.0 (the number of output filters based on the Width Multiplier\color{black})
\end{itemize}
Indeed, for large output depths, the number of MACs almost scales linearly with the Width Multiplier. For large output depths, the size of the filters (3×3) becomes negligible in comparison to the large number of filters, which is proportional to the Width Multiplier\color{black}.

\subsection{ Semantic Segmenter} \label{sec:segmenter}
\noindent{\textbf{1x1 Convolutions as Pixel-wise Classifiers.}}
A 1 × 1 convolutional layer processes the output derived from the MobileNet feature extractor, serving as the classifier. In the context of convolutional neural networks, 1 × 1 convolutions as pixel-wise classifiers refer to the use of 1 × 1 convolution layers to perform pixel-wise classification. 

In image classification tasks, a fully connected layer is often used to process all pixels in an input feature map for prediction. However, for pixel-wise classification, where each pixel is classified individually, the approach needs modification. In this case, each pixel slice of $c_{\text{i}}$ connects with all n elements of the output vector. 

When using a convolution layer with $n$ filters, the filter height and width are set to 1 ($h_{\text{f}} = w_{\text{f}} = 1$) to achieve pointwise convolution.
\color{black}
Using n filters, a volume with identical spatial dimensions as the input (1 × 1) and a depth of n is created. This essentially forms another one-pixel column across the depth of n. As a result, both the input and output dimensions for the fully connected layer and the 1 × 1 (pointwise) convolution layer become identical.

The computations in the 1x1 convolution layer are equivalent to those in a fully connected layer. Each output pixel is generated by the element-wise multiplication of a filter’s weights with the input column, both having dimensions [1 × 1 × $c_{\text{i}}$], resulting in $c_{\text{i}}$ direct, learnable connections from each input pixel to its corresponding output pixel.

With knowledge of the presence of n output pixels, the total count of learnable weights in a 1x1 convolution layer can be calculated as $c_{\text{i}}$ × n, similar to a fully connected layer. However, unlike fully connected layers, convolutional layers are flexible with varying spatial input sizes. For example, when applying pixel-wise convolution to an image with dimensions W × H and N channels ($c_{\text{i}}$ = N), a 1 × 1 × N filter is used, and using N filters results in input and output volumes with the same dimensions. By setting the number of filters, $n_{\text{f}}$, to n, n output feature maps are obtained, one per class.

\begin{figure}[!t]
  \centering
  \includegraphics[width=1.0\linewidth]{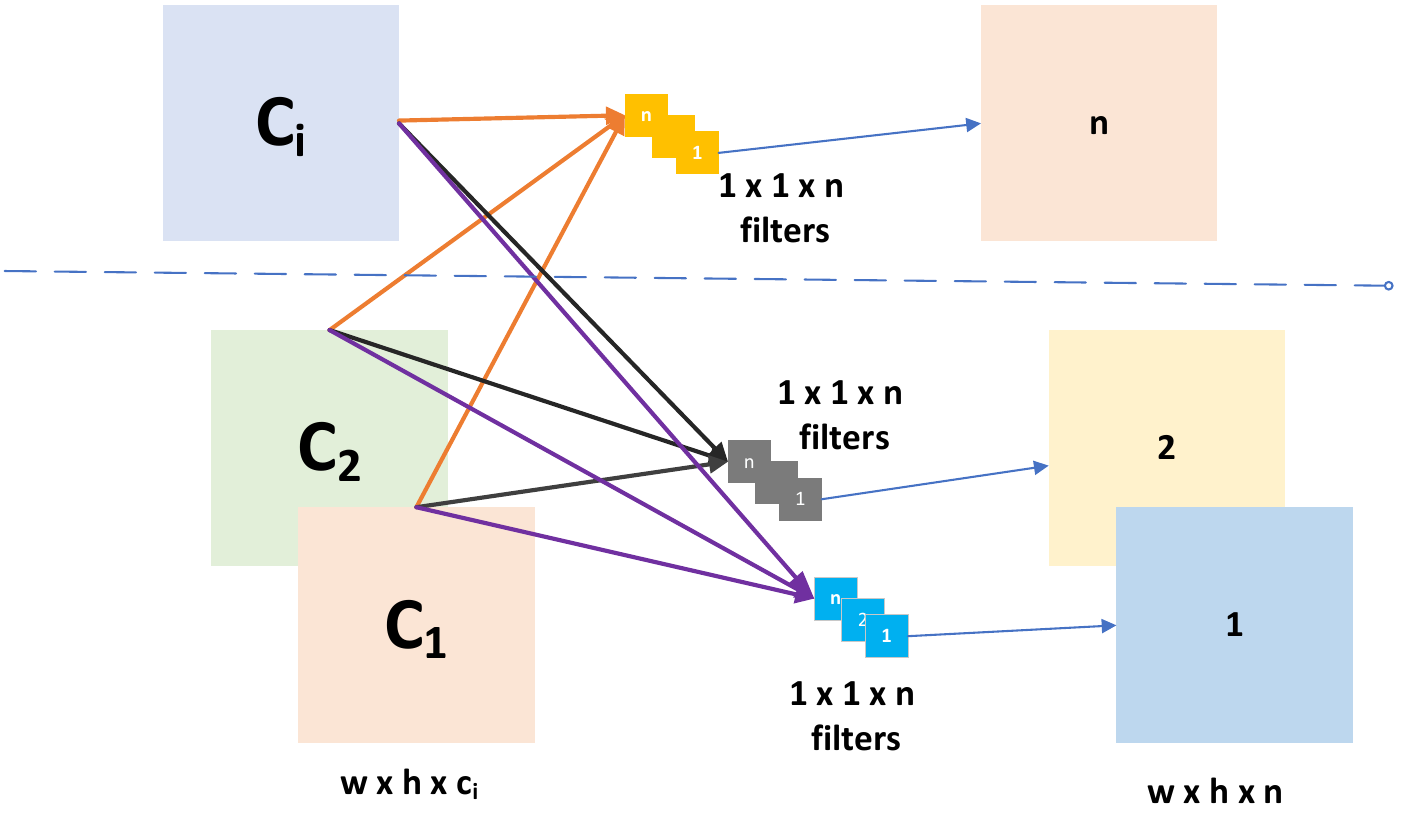}
  \caption{
1x1 convolutions as pixel-wise classifiers.
  }
  \label{fig:pixel-wise}
\vspace{-4mm}
\end{figure}

\noindent{\textbf{Transposed Convolution for Upsampling.}}
The layer used for upsampling in a Fully Convolutional Network is the Transposed Convolution~\cite{7298965}, which acts as a 'brush' to distribute input pixel values, allowing us to upsample feature maps. By setting stride s=1 and applying 'full' padding p=(f-1)/2 (based on filter size), the spatial dimensions of the input and output feature maps remain the same. Transposed Convolution moves a filter window over an output volume initialized with zeros, taking a single input pixel, multiplying it with all filter weights, and adding the result to the corresponding elements in the output. Zero-padding in Transposed Convolution adds pixels to the output feature map.

Given an input of 2 × 2 pixels (i=2) with only one channel, and a transposed convolution kernel of 3 × 3 pixels (k=3) without any zero padding (p=0) and a stride of s=1, we can calculate the output dimensions using the side length o:
\begin{equation}
\begin{aligned}
o &= (i-1) \times s + k - 2p.
\label{eq:transpose}
\end{aligned}
\end{equation}
For the specific settings mentioned, the equation simplifies to 
\[\begin{aligned}
o= (2-1) \times 1 + 3 - 2 \times 0
&= 4.
\end{aligned}
\]

Therefore, the output dimensions for this scenario would be 4 × 4 pixels.

To upscale an input feature map by a factor of x, it is crucial to ensure overlapping 'brush strokes' in all Transposed Convolution operations. This is achieved by setting the stride (s) to half the kernel size (k), ensuring each window overlaps with its neighbors. This overlap preserves spatial correlations between adjacent regions, enhancing model performance.
To achieve an output size of o = x, a shorthand trick is utilized by setting s = x, k = 2s = 2x, and p = s/2 = x/2. Eq.~\ref{eq:transpose} allows us to derive the following expression:
\[
\begin{aligned}
o &= s \times (i - 1) + k - 2 \times p \\
&= x \times (i - 1) + 2 \times x - 2 \times \left(\frac{x}{2}\right) \\
&= x
\end{aligned}
\]

In parameter space, we can define a function $\theta(x)$ that represents the upscaling operation for an input by a target factor $x$ as:

\[
\theta(x) : \mathbb{R} \rightarrow \mathbb{N}_0^3,
\]
where
\begin{equation}
\begin{aligned}
\theta(x) = 
\begin{pmatrix}
s \\
k \\
p \\
\end{pmatrix}
=
\begin{pmatrix}
x \\
2x \\
\frac{x}{2} \\
\end{pmatrix},
\label{eq:scailng}
\end{aligned}
\end{equation}
where $s$ represents the stride, $k$ denotes the kernel size, and $p$ corresponds to the zero padding parameters of the Transposed Convolution layer.
$\theta(x)$ is a vector $(s, k, p)$ representing these parameters, where $x$ is the desired factor for upsampling.

Using this configuration, the overlapping sliding window ensures effective pixel distribution during upsampling. Similar to a marker's stroke, if all weights in a Transposed Convolution kernel are identical, the output pixels within the window will have uniform intensity based on the corresponding input pixel. This setup enables accurate and efficient upsampling, essential for generating high-resolution segmentation maps in Fully Convolutional Networks.

From the fully convolutional MobileNet base topology, it has been determined that downsampling occurs by a factor of 32 both horizontally and vertically. Consequently, to achieve a classification granularity of one distinct class per pixel, the same factor must perform an upsampling operation. This indicates that Eq.~\ref{eq:scailng} is utilized for the stride (s), kernel size (k), and padding (p) to achieve this. Based on this, we can conclude that using large square kernels with an edge length of k=64 is preferable. Instead of emulating a marker pen, a ``real" brush shape is desired, where more color is applied to the central regions compared to the surroundings. In fact, the upsampling strategy is commonly referred to as upsampling with bilinear interpolation~\cite{prashanth2009image}. The algorithms of the bilinear distribution to initialize the weights for transposed convolution are given in Algorithm~\ref{alg:bilinear_kernels}.

\begin{algorithm}[tb]
\caption{Create Bilinear Kernels}
\label{alg:bilinear_kernels}
\begin{algorithmic}[1]
\Function{CreateBilinearKernels}{$num\_classes$, $kernel\_dims = [64, 64]$}
    \State \textbf{function} \textsc{find\_params}($size$)
        \State $\text{divisor} \gets \left\lfloor \frac{\text{size} + 1}{2} \right\rfloor$
        \If{$size \% 2 == 0$}
            \State $center \gets divisor - 0.5$
        \Else
            \State $center \gets divisor - 1$
        \EndIf
        \State \textbf{return} $divisor$, $center$
    \State \textbf{end function}

    \State $height, width \gets kernel\_dims$
    \State $divisor\_height, center\_height \gets \textsc{find\_params}(height)$
    \State $divisor\_width, center\_width \gets \textsc{find\_params}(width)$
    \State $grid \gets \text{ogrid}[:height, :width]$
    \State $filter\_coefficients \gets (1.0 - |grid[0] - center\_height| / divisor\_height)$ 
           $\times (1.0 - |grid[1] - center\_width| / divisor\_width)$
    
    \State $one\_plane \gets \text{array}(filter\_coefficients,$ 
           $\text{dtype=float32})$
           
    \State $all\_planes \gets \text{zeros}((height, width,$ 
           $num\_classes, num\_classes), \text{dtype=float32})$

    \For{$i \gets 1$ \textbf{to} $num\_classes$}
        \For{$j \gets 1$ \textbf{to} $num\_classes$}
            \State $all\_planes[:, :, i, j] \gets one\_plane$
        \EndFor
    \EndFor
    \State $coefficients \gets all\_planes$
    \State \textbf{return} $coefficients$
\EndFunction
\end{algorithmic}
\end{algorithm}

\noindent{\textbf{Key Steps Explanation of Algorithm~\ref{alg:bilinear_kernels} }}
\begin{enumerate}
    \item \textbf{Initialization:}
The algorithm takes two parameters, \texttt{num\_classes} and \texttt{kernel\_dims}, with \texttt{kernel\_dims} defaulting to \([64, 64]\).\\
        num\_classes is the number of classes for classification.\\
        kernel\_dims: is the dimensions of the transposed convolution kernel, and the default is [64, 64].
    \item \textbf{Calculate Center and Divisor:}
The \texttt{find\_params} function calculates the center and divisor based on the given kernel size.
    \item \textbf{Generate Bilinear Interpolation Kernel:}
Create a 2D grid and compute the bilinear interpolation coefficients, which are stored in \texttt{one\_plane}.
    \item \textbf{Initialize All Kernel Weights:}
Create a zero array of shape \([ \text{height} \times \text{width} \times \text{num\_classes} \times \text{num\_classes} ]\) and copy \texttt{one\_plane} to every position in the last two dimensions.
    \item \textbf{Return Result:}
Return \texttt{coefficients}, an array with the bilinear filter kernel weights.
\end{enumerate}

\begin{figure}[t]
\vspace{-0mm}
  \centering
  \includegraphics[width=1.0\linewidth]{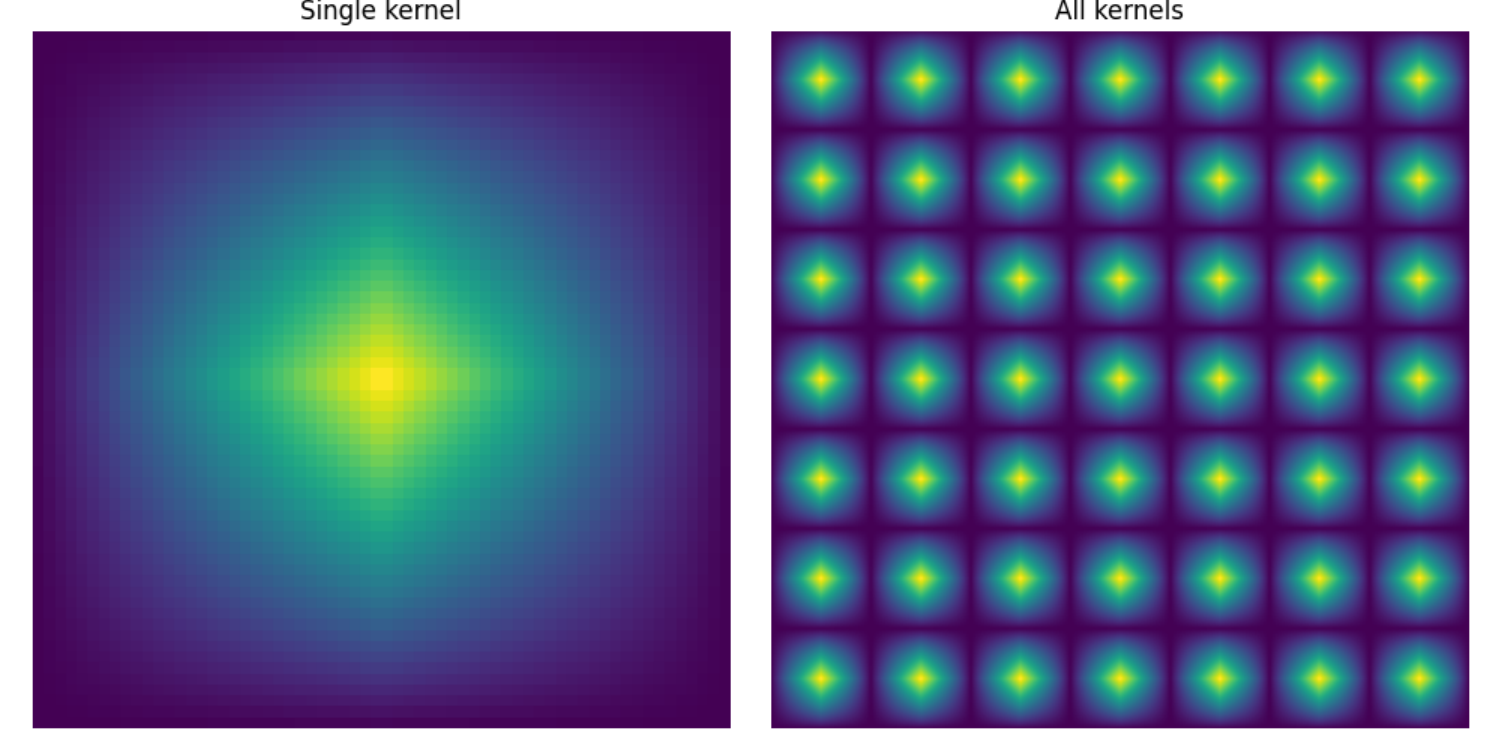}
  \caption{
Use a bilinear distribution as the weights initializer for the Transposed Convolution kernels in our architecture. The Left shows a single bilinear convolution kernel (64x64). This kernel is used for upsampling operations. Right thumbnails showing all generated bilinear convolution kernels.  }
  \label{fig:billnear}
\vspace{-2mm}
\end{figure}  

In Figure~\ref{fig:billnear}, we show a visual representation using a 64x64 bilinear convolution kernel and seven classification categories. The left plot illustrates a single bilinear kernel, commonly used for upsampling, with values highest at the center and decreasing towards the edges, facilitating smooth transitions. The right plot shows thumbnails of all generated kernels, each with the same distribution but arranged differently for different classes. This helps confirm the consistency of kernel generation. These bilinear kernels enhance upsampling in tasks like image segmentation by restoring low-resolution features to higher resolution, improving pixel-level accuracy.

\subsection{Semantic Segmenter with an Adjustable Feature Extractor}
By combining the adjustable feature extractor with the semantic segmenter, our approach achieves enhanced accuracy and flexibility in image segmentation. The adjustable feature extractor allows us to adaptively extract features that are most relevant to the segmentation task at hand, while the semantic segmenter provides a comprehensive understanding of the image content for precise segmentation. Algorithm~\ref{alg:add_fcn_head} is designed to adaptively extract relevant features from images based on the specific requirements of the segmentation task. Adjusting its parameters, such as kernel sizes, width multiplier, and depth, we can effectively capture and emphasize important visual patterns in the input images.

\begin{algorithm}[tb]
\caption{Adjustable Semantic Segmentation Network}
\label{alg:add_fcn_head}
\begin{algorithmic}[1]
\Function{AddFCNHead}{$net$, $input\_shape$, $is\_training$, 
       \newline \hspace{1.8cm}$CLASSIFIER\_KERNEL$, $WIDTH\_MULTIPLIER$,
       \newline \hspace{1.8cm}$CLASSIFIER\_DEPTH$, $NUMBER\_CLASSES$}
    
    \State $conv1\_separable \gets \text{SeparableConv}(input,$ 
           \newline \hspace{2cm}$CLASSIFIER\_KERNEL, DEPTH\_MULTIPLIER)$
    \Comment{Depthwise separable convolution block}
    
    \State $conv1\_pointwise \gets \text{PointwiseConv}(net,$
           \newline \hspace{2cm}$CLASSIFIER\_DEPTH)$
    \Comment{1x1 pointwise convolution}
    
    \State $conv2\_classifier \gets \text{Conv1x1}(NUMBER\_CLASSES)$
    \Comment{Classifier to output num\_classes feature maps}
    
    \State $current\_shape \gets \{net.shape\}$
    \State $factor\_height \gets input\_shape[1] / current\_shape[1]$
    \State $factor\_width \gets input\_shape[2] / current\_shape[2]$
    \State $upsampling\_factor \gets 32$
    \State $upsample\_kernel\_size \gets 64$
    
    \If{$is\_training$}
        \State $filter\_coefficients \gets$
               \newline \hspace{1cm}$\text{Create\_bilinear\_kernels}(NUMBER\_CLASSES,$
               \newline \hspace{1.7cm}$upsample\_kernel\_size)$
        \State $\text{InitializeWeights}(filter\_coefficients)$
    \Else
        \State $filter\_coefficients \gets [0]$
    \EndIf
    
    \State $upsample\_stride \gets 32$
    \State $upsampler \gets \text{TransposedConv}(conv2\_classifier,$
           \newline \hspace{2cm}$NUMBER\_CLASSES, kernel\_size=upsample\_kernel\_size,$
           \newline \hspace{2cm}$stride=upsample\_stride)$
    \Comment{Upsampling to original resolution}
    
    \State \Return $upsampler$
\EndFunction
\end{algorithmic}
\end{algorithm}

\noindent{\textbf{Algorithm~\ref{alg:add_fcn_head} Explanation}}
\vspace{-5mm}
\subsection*{Input:}
\begin{enumerate}
    \item \textbf{net}: The output tensor of the base network.
    \item \textbf{input\_shape}: Shape of the input image tensor.
    \item \textbf{is\_training}: Boolean indicating if the network is in training mode.
    \item \textbf{WIDTH\_MULTIPLIER}: 
    This parameter controls the number of filters per input channel during the depth-wise operation. Adjusting \texttt{DEPTH\_MULTIPLIER} regulates the output channels per input channel, allowing us to assess its impact on the FCN architecture's performance.

    \item \textbf{CLASSIFIER\_KERNEL}: 
    This parameter defines the kernel size in the Separable Convolution layer of the FCN head. Varying the kernel size helps evaluate its impact on segmentation performance, allowing us to identify the optimal size for accurate and detailed results.
    
    \item \textbf{CLASSIFIER\_DEPTH}: 
    This parameter controls the depth of the first pointwise convolution layer in the FCN head, affecting the network's feature extraction capabilities. Empirical evaluation helps identify the optimal depth for improved segmentation performance.
    
    \item \textbf{NUMBER\_CLASSES}: 
    This parameter specifies the number of classes to be segmented based on the use-case scenario. Exploring different scenarios helps assess how varying \texttt{NUMBER\_CLASSES} affects the FCN's segmentation accuracy.

\end{enumerate}
 \vspace{-6mm}
\subsection*{Steps:}
\begin{enumerate}
    \item \textbf{Depth-wise Convolution Block:}
    \begin{enumerate}
        \item Apply a depth-wise separable convolution with the specified kernel size and batch normalization.
        \item Follow with a point-wise convolution to adjust the depth to the specified number of filters.
    \end{enumerate}
   
    \item \textbf{Classifier Convolution Layer:}
    \begin{enumerate}
        \item Apply a 1x1 convolution to generate the final class scores.
    \end{enumerate}
    \item \textbf{Transposed Convolution Layer (Upsampling):}
    \begin{enumerate}
        \item Ensure that the upsampling factor is correct (that is, the input shape dimensions divided by the current shape dimensions equals 32).
        \item Initialize the filter coefficients for the transposed convolution layer.
        \item Apply the transposed convolution to upsample the feature map back to the original input image resolution.
    \end{enumerate}
\end{enumerate}

\subsection{Exploring the Parameter Space} 
\label{sec:segmenter2}

Based on Table~\ref{tbl:three_basic}, it is noted that existing methods struggle with the specific resource requirements, frame rate, and accuracy of target scenarios. Tailored optimization strategies are needed to address these challenges effectively:


\subsubsection{{\textbf{Diverse Scenario-Specific Requirements}}}
Different scenarios demand varying numbers of classes, camera counts, frame rates, and accuracy levels. For example, parking and urban environments both need to process 7 classes but have different computational budgets and frame rates. Existing methods may not adapt well to these specific constraints, leading to either suboptimal performance or excessive resource consumption.

\subsubsection{\textbf{Computational Budget Constraints}}
Scenarios have different computational budgets, such as 70, 120, and 300 Giga operations per second. It is crucial to meet these constraints while maintaining performance. Many methods assume ample resources and do not optimize for constrained budgets, resulting in performance issues.

\subsubsection{\textbf{Frame Rate Requirements}}
Different scenarios require varying frame rates, such as 15 fps versus 30 fps. This impacts real-time processing and computational demands. Existing methods may not effectively address the impact of frame rate on processing speed, leading to inefficiencies in high frame rate scenarios.

\subsubsection{\textbf{Accuracy Variability}}
Accuracy requirements also vary across scenarios. Urban environments demand high accuracy, while parking lots might only require medium accuracy. Current methods might not adjust well to these varying accuracy needs, affecting performance in different applications.

\begin{algorithm}[tb]
\caption{Bayesian Optimization for Hyper-parameter Search}
\label{alg:nas}
\begin{algorithmic}[1]
\Function{BayesianHyperparamSearch}{$n\_cameras$, $fps\_per\_camera$,
       \newline \hspace{2cm}$num\_classes$, $computational\_budget$,
       \newline \hspace{2cm}$max\_iterations$}
    
    \State Initialize $WIDTH\_MULTIPLIER \gets [0.25, 0.5, 0.75, 1.0, 1.25]$
    \State Initialize $CLASSIFIER\_KERNEL \gets [3, 5, 7, 9, 11]$
    \State Initialize $CLASSIFIER\_DEPTH \gets [512, 1024, 1536, 2048]$
    \State Initialize $block\_specs\_list \gets [block\_specs_1, block\_specs_2]$
    
    \State $images\_per\_second \gets n\_cameras \times fps\_per\_camera$
    \State $min\_difference \gets \infty$
    \State $closest\_configuration \gets None$
    
    \State Initialize surrogate model \Comment{Gaussian Process}
    
    \For{$iteration \gets 1$ \textbf{to} $max\_iterations$}
        \State Select $m, d, k, block\_specs\_index$ using
               \newline \hspace{1cm}acquisition function \Comment{Based on surrogate model}
        
        \State $block\_specs \gets block\_specs\_list[block\_specs\_index]$
        \State Initialize $model$ with $block\_specs$, $num\_classes$,
               \newline \hspace{2cm}$depth\_multiplier=m$
        \State Set $model$ classifier with $kernel=k$ and $depth=d$
        
        \State $macs \gets get\_megaops(model)$
        \State $gigaops\_per\_second \gets$
               \newline \hspace{1cm}$\frac{images\_per\_second \times macs}{1000}$
        
        \State Update surrogate model with $(m, d, k, block\_specs\_index$,
               \newline \hspace{2cm}$gigaops\_per\_second)$
        
        \If{$gigaops\_per\_second \leq computational\_budget$ \textbf{and}
           \newline \hspace{1cm}$(computational\_budget - gigaops\_per\_second) < min\_difference$}
            \State $min\_difference \gets computational\_budget - gigaops\_per\_second$
            \State Update $closest\_configuration$ with current
                   \newline \hspace{1.5cm}hyper-parameters
        \EndIf
    \EndFor
    
    \State \Return $closest\_configuration$
\EndFunction
\end{algorithmic}
\end{algorithm}

\color{black}
Bayesian Optimization is a probabilistic model-based optimization technique particularly effective for optimizing functions that are expensive to evaluate. It constructs a surrogate model of the objective function, typically a Gaussian Process, to predict the performance of different hyper-parameter configurations based on past evaluations.
The process works as follows: 

Initially, a surrogate model is created using prior evaluations. This model estimates the objective function's value for unexplored hyper-parameter configurations.

Next, an acquisition function is used to determine the next set of hyper-parameters to evaluate. The acquisition function balances exploration, by testing new areas of the hyper-parameter space, and exploitation, by refining known promising areas.

Finally, the algorithm iteratively updates the surrogate model with new evaluations. This improves the accuracy of the model’s predictions over time and effectively guides the search for optimal hyper-parameters.

In addressing the tasks specified in Table~\ref{tbl:three_basic}, we applied Bayesian optimization to adjust the model's hyperparameters for the DRIVE PX 2 platform. 
While Bayesian Optimization has been applied in Neural Architecture Search (NAS), our use of \textbf{Bayesian Optimization with surrogate modeling} introduces a more efficient way to explore hyperparameter spaces under tight computational budgets (e.g., DRIVE PX 2). 

Unlike previous approaches that rely on exhaustive or grid searches, our method utilizes a Gaussian Process-based surrogate model to predict the performance of untested configurations, significantly reducing the need for exhaustive exploration. By leveraging Bayesian Optimization, we intelligently navigate the hyperparameter space, optimizing model complexity within computational constraints. This allows for a more efficient search compared to traditional NAS techniques, which often lack real-time adaptability or surrogate modeling.

The Algorithm~\ref{alg:nas} process is as follows:

\begin{enumerate}
    \item \textbf{Initialization (Steps 1-6)}:
    \begin{itemize}[leftmargin=*]
        \item The input parameters include the number of cameras, frames per second (\textit{fps}) per camera, the number of classes, the computational budget, and the maximum number of iterations.
        \item Predefined lists for \textit{WIDTH\_MULTIPLIER}, \textit{CLASSIFIER\_KERNEL}, and \textit{CLASSIFIER\_DEPTH} are initialized. Additionally, a list of block specifications is set.
        \item The algorithm calculates the number of images processed per second and initializes variables to track the best configuration relative to the computational budget.
    \end{itemize}
    
    \item \textbf{Surrogate Model Initialization (Step 8)}:
    \begin{itemize}[leftmargin=*]
        \item A \textbf{Gaussian Process} is initialized as the surrogate model. This model assists in predicting the performance of various hyper-parameter configurations by leveraging prior evaluations.
    \end{itemize}
    
    \item \textbf{Iterative Optimization (Steps 10-21)}:
    \begin{itemize}[leftmargin=*]
        \item A loop is performed for a maximum number of iterations.
        \item At each iteration, the \textbf{Acquisition Function} selects hyper-parameters such as the width multiplier, classifier depth, kernel size, and block specifications based on predictions made by the surrogate model.
        \item A model is then initialized using the selected hyper-parameters, and the \textit{Multiply-Accumulate Operations (MACs)} are calculated.
        \item The model's computational efficiency is evaluated by computing its \textit{giga-operations per second}.
    \end{itemize}
    
    \item \textbf{Model Evaluation (Steps 18-21)}:
    \begin{itemize}[leftmargin=*]
        \item The surrogate model is updated with the new hyper-parameters and their corresponding computational results.
        \item If the model’s computational efficiency is within the budget and improves upon previous configurations, the current configuration is stored as the best.
    \end{itemize}
    
    \item \textbf{Output (Step 23)}:
    \begin{itemize}[leftmargin=*]
        \item After all iterations, the algorithm returns the configuration that best meets the computational budget while maximizing performance.
    \end{itemize}
\end{enumerate}
\color{black}
\noindent{\textbf{Key Points Addressed by the Algorithm}}
\begin{itemize}[leftmargin=*]
  \item  \underline{Bayesian Optimization:} The algorithm uses Bayesian optimization (line 11) to select hyper-parameters, constructing a probabilistic model of the objective function and iteratively refining choices based on previous results, offering a more efficient search than exhaustive methods.
  \item  \underline{Targeted Search:} The search focuses on configurations likely to meet specific constraints, such as computational budgets (line 18), optimizing hardware and task requirements without evaluating all possibilities.
  \item  \underline{Branch Pruning:} The conditional check (line 18) acts as branch pruning, discarding configurations that don't meet budget constraints early in the process, updating only when improvements are found.
  \item   \underline{Iterative Improvement:} The iterative nature of Bayesian optimization narrows the search to promising areas, continuously improving hyper-parameter selection without exploring all options.
\end{itemize}

\color{black}
\noindent{\textbf{Comparison with Previous NAS Methods}}

Bayesian Optimization (BO) methods have been widely applied in Neural Architecture Search (NAS) literature~\cite{10.5555/3326943.3327130,white2021bananas,liu2024efficient} due to their ability to efficiently navigate complex search spaces and balance exploration and exploitation. Our approach shares some fundamental aspects with these methods but also exhibits distinct differences:

First, in terms of surrogate model flexibility, while most BO-based NAS methods use either a Gaussian Process (GP)\cite{10.5555/3326943.3327130} which focuses heavily on discrete architecture choices\cite{pmlr-v80-pham18a}, our approach provides additional flexibility. Specifically, our method allows more customization in network design, including block specifications, kernel sizes, and classifier depth, allowing better adaptation to specific tasks.

Second, regarding the acquisition function, standard NAS frameworks often employ functions like Upper Confidence Bound (UCB)~\cite{white2021bananas}, but they don't always consider computational constraints. Our method introduces a custom acquisition function tailored specifically to our architecture's requirements. This ensures a more effective balance between exploration and exploitation while factoring in hardware's computational limitations.

Third, with respect to optimization for computational efficiency, many NAS methods focus primarily on optimizing architectures for accuracy~\cite{liu2024efficient}, without much consideration for efficiency. In contrast, our approach iteratively optimizes architectures while explicitly considering computational budgets and hardware constraints. By incorporating efficiency metrics like Giga operations per second (GigaOps), our method is well-suited for resource-constrained environments.

Finally, in terms of search space and flexibility, rather than using a fixed architecture search space, our approach allows for greater adaptability in network design, particularly with respect to block specifications, kernel sizes, and classifier depth. This adaptability enables better optimization of performance for specific tasks

Our method's key innovations—awareness of computational constraints, flexibility in surrogate modeling, and real-world applicability—set it apart from traditional NAS techniques. This gives our approach a distinct advantage in scenarios where both efficiency and hardware performance are as important as accuracy. We efficiently adjusted the model hyperparameters to meet the specific requirements of the DRIVE PX 2 platform across various tasks, ensuring optimal use of computational resources while maximizing model performance within the given budget and accuracy targets.

\color{black} 



%% file: sections/6-exp-results.tex
\section{Experiments}
In this section, we introduce the experiments and the results to verify the effectiveness and efficiency of our method.
\subsection{Experiment Setup}
Our experiment is implemented in NVIDIA Deep Learning GPU Training System (DIGITS), the experimental hardware is NVIDIA DRIVE PX 2 which connects up to 12 cameras and uses the CamVid~\cite{BrostowSFC:ECCV08} and Cityscapes~\cite{cordts2016cityscapes} dataset, the model is trained and tested using the NVIDIA DIGITS. The model backbone is MobileNetV4~\cite{qin2024mobilenetv4}. 

\subsection{Exploring Hyper-parameters}
This exploration seeks to provide a comprehensive understanding of the possibilities and variations achievable by manipulating these parameters. By thoroughly examining the design space, the automatic parameter search algorithm can gain insights into how parameter modifications impact the model's performance. The goal is to identify optimal configurations that meet the strict requirement of the available budget in Giga operations per second. \\
\noindent{\textbf{Rural Scenario.}} 
According to the requirements specified for the Rural Scenario in Table~\ref{tbl:three_basic}, we will configure the following parameters: 2 classes, 1 camera, each capturing 30 frames per second, and a computational budget of 120 Giga operations per second as the Algorithm~\ref{alg:nas} input. Next, we employ Algorithm~\ref{alg:nas} to search for the optimal value {WIDTH\_MULTIPLIER}, {CLASSIFIER\_KERNEL}, {CLASSIFIER\_DEPTH} that satisfies the computational budget constraint. Through this search, we obtain the following values: mode selects, {WIDTH\_MULTIPLIER} = 1.25, {CLASSIFIER\_DEPTH} = 2048, and {CLASSIFIER\_KERNEL} = 7. These values result in a computational requirement of 117.74 Giga operations per second, which maximally satisfies the budget of 120 Giga operations per second.

\noindent{\textbf{Urban Scenario.}} 
According to the requirements specified for the Urban scenario in Table~\ref{tbl:three_basic}, we configure the following parameters: 7 classes, 4 cameras, each capturing 30 frames per second, and a computational budget of 300 Giga operations per second as the Algorithm~\ref{alg:nas} input. Next, we employ Algorithm~\ref{alg:nas} to search for the optimal value {WIDTH\_MULTIPLIER}, {CLASSIFIER\_KERNEL}, {CLASSIFIER\_DEPTH} that satisfies the computational budget constraint. Through this search, we obtain the 
following values: {WIDTH\_MULTIPLIER} = 0.25, {CLASSIFIER\_DEPTH} = 512, and {CLASSIFIER\_KERNEL} = 11. These values result in a computational requirement of 299.64 Giga operations per second, which maximally satisfies the budget of 300 Giga operations per second.

\noindent{\textbf{Parking Scenario.}} 
According to the requirements specified for the Parking scenario in Table~\ref{tbl:three_basic}, we will configure 4 cameras, each capturing 15 frames per second, with a computational budget of 70 Giga operations per second as input for Algorithm~\ref{alg:nas}. Through this algorithm, we obtain the following values: WIDTH\_MULTIPLIER = 1.25, CLASSIFIER\_DEPTH = 2048, and CLASSIFIER\_KERNEL = 3. These parameters result in a computational requirement of 69.78 GOPS, which satisfies the budget of 70 GOPS.

\subsection{Experimental Results }

\begin{table}[tb!]
    \vspace{0mm}
    \centering
    \caption{COMPARISONS WITH REAL-TIME METHODS ON THE CAMVID. } 
    \resizebox{\linewidth}{!}{
    \begin{tabular}{l|c|c|c|cc}
     \toprule
        \textbf{Method} &\textbf{gFLOP} & \textbf{FPS} &  \textbf{Test IOU} &  \textbf{GPU}   \\
        \midrule
        \multirow{1}{*}{DfaNet A~\cite{li2019dfanet} }   &{1.70} &{160} &59.3 &Titan X\\
        \multirow{1}{*}{BiseNet~\cite{yu2018bisenet} }   &{2.90} &{-} &65.6 &1080ti\\

         \multirow{1}{*}{ENet~\cite{paszke2016enet} }   &{3.83} &{21.1} &51.3 &NV TX1\\
        
        \multirow{1}{*}{EfficitNet-B0 +IFVD~\cite{koonce2021efficientnet} }   &{7.96} &{-} &64.4 &Titan X\\
        \multirow{1}{*}{HySeg-S~\cite{zhang2021dcnas} }   &{17.00} &{38.0} &78.4 &1080ti\\
        \multirow{1}{*}{RTFormer-Slim~\cite{wang2022rtformer}  }   &{17.50} &{190.7} &81.4 &RTX 2080Ti\\
        \multirow{1}{*}{LinkNet~\cite{chaurasia2017linknet} }   &{21.20} &{9.3} &68.3 &NV TX1\\
        \multirow{1}{*}{ICNet~\cite{zhao2018icnet} }   &{28.30} &{27.8} &75.42 &Titan X\\
        \multirow{1}{*}{SegNet+CAP-30\%~\cite{badrinarayanan2017segnet} }   &{30.01} &{-} &56.37 &Titan X\\
       \multirow{1}{*}{DDRNet~\cite{hong2021deep} }   &{36.00} &{180} &74.7 &G1660ti\\
        
        \multirow{1}{*}{SegNet~\cite{badrinarayanan2017segnet} }   &{286.00} &{1.3} &55.6 &NV TX1\\
     
        \midrule
        \rowcolor{light-gray} 
        \multirow{1}{*}{TSLA-Large}   &{1.98} &{77.30} &78.6 &Titan X\\
        \rowcolor{light-gray} 
        \multirow{1}{*}{TSLA-Large}   &{2.02} &{61.20} &78.6 &Pascal d\\
        \rowcolor{light-gray} 
         \multirow{1}{*}{TSLA-Medium}   &{1.24} &{85.0} &72.5 &Titan X\\
        \rowcolor{light-gray} 
         \multirow{1}{*}{TSLA-Medium}   &{1.35} &{92.0} &72.5 &Pascal d\\
        \rowcolor{light-gray} 
         \multirow{1}{*}{TSLA-Small}   &{0.52} &{92.2} &70.3 &Titan X\\
         \rowcolor{light-gray} 
         \multirow{1}{*}{TSLA-Small}   &{0.55} &{108.6} &70.3 &Pascal d\\
        \bottomrule
    \end{tabular}
   }
   
   \label{tbl:camvid}
\end{table}

\begin{table}[tb!]
    \vspace{0mm}
    \centering
    \caption{COMPARISONS WITH REAL-TIME METHODS ON THE CITYSCAPES DATASET. } 
    \resizebox{\linewidth}{!}{
    \begin{tabular}{l|c|c|c|cc}
     \toprule
        \textbf{Method} &\textbf{GFLOPs} & \textbf{FPS} &  \textbf{Test IOU} &  \textbf{GPU}   \\
        \midrule

        \multirow{1}{*}{DfaNet A~\cite{li2019dfanet} }   &{2.10} &{120} &67.10 &Titan X\\
        
        \multirow{1}{*}{BiseNet~\cite{yu2018bisenet} }   &{2.90} &{-} &68.40 &1080ti\\
        \multirow{1}{*}{ENet~\cite{paszke2016enet} }   &{3.83} &{21.6} &58.3 &NV TX1\\ 
        \multirow{1}{*}{RAFNet~\cite{chen2023rafnet}  }   &{4.15} &{212} &70.5&RTX A6000\\
        \multirow{1}{*}{HySeg-m~\cite{zhang2021dcnas} }   &{7.50} &{36.90} &75.80 &1080ti\\
        \multirow{1}{*}{EfficitNet-B0 +IFVD~\cite{koonce2021efficientnet} }   &{7.96} &{-} &62.52 &Titan X\\
        \multirow{1}{*}{SegFormer~\cite{xie2021segformer} } &{12.40} &{3.0} &83.8 &-\\
        \multirow{1}{*}{HRViT-b1+~\cite{gu2022multi}  }   &{14.10} &{-} &81.36 &-\\
        \multirow{1}{*}{ResNet101+TEM~\cite{michieli2022edge}  }   &{17.48} &{-} &80.9 &Titan X\\
        \multirow{1}{*}{RTFormer-Slim~\cite{wang2022rtformer}  }   &{17.50} &{110.0} &76.3 &RTX 2080Ti\\
        \multirow{1}{*}{ICNet+CAP-60\%~\cite{li2020semantic} }   &{21.16} &{-} &62.38 &Titan X\\
         \multirow{1}{*}{FastrSeg~\cite{chen2019fasterseg}  }   &{28.20} &{163.9} &69.5 &1080ti\\
       
        \multirow{1}{*}{ICNet~\cite{zhao2018icnet} }   &{28.30} &{27.8} &75.42 &Titan X\\
       
        \multirow{1}{*}{DRNet-23-slim~\cite{hong2021deep} }   &{36.30} &{101.6} &77.4 &2080ti\\
       
        \multirow{1}{*}{PIDNet-S-Simple~\cite{sun2020pidnet} }   &{46.30} &{100.8} &78.2 &1080ti\\
          \multirow{1}{*}{MiT-B0+ DynaSegFor.~\cite{he2021cap}  }   &{46.70} &{-} &74.35 &V100\\
       
        \multirow{1}{*}{BisNetR18+m0~\cite{bai2022dynamically} }   &{75.48} &{-} &72.45 &-\\
       \multirow{1}{*}{MobileNet V3-Large~\cite{howard2019searching} } &{2.48} &{63.16} &76.6 &TPU Pod\\
       
       \multirow{1}{*}{MobileNet V3-Small~\cite{howard2019searching} } &{0.74} &{76.31} &68.38 &TPU Pod\\

        \midrule
        \rowcolor{light-gray} 
        \multirow{1}{*}{TSLA-Large}   &{1.98} &{76.15} &78.80 &Titan X\\
        \rowcolor{light-gray} 
        \multirow{1}{*}{TSLA-Large}   &{2.17} &{60.56} &78.75 &Pascal d\\
        \rowcolor{light-gray} 
         \multirow{1}{*}{TSLA-Medium}   &{1.24} &{81.17} &72.60 &Titan X\\
         \rowcolor{light-gray} 
         \multirow{1}{*}{TSLA-Medium}   &{1.35} &{72.39} &72.60 &Pascal d\\
        \rowcolor{light-gray} 
         \multirow{1}{*}{TSLA-Small}   &{0.52} &{96.52} &67.40 &Titan X\\
        \rowcolor{light-gray} 
         \multirow{1}{*}{TSLA-Small}   &{0.63} &{118.23} &67.40 &Pascal d\\
        \bottomrule
    \end{tabular}
   }
   
   \label{tbl:city}
\end{table}
According to the information provided in Table~\ref{tbl:camvid} and Table~\ref{tbl:city}, cutting-edge models demonstrate high Intersection over Union (IOU) scores, but they are typically implemented on high-end NVIDIA GPU servers with significantly high Floating Point Operations Per Second (FLOPS) requirements. Consequently, the widespread adoption of these models in driverless cars faces obstacles. 
It should be noted that even after pruning, certain models such as DDRNet36-slim~\cite{hong2021deep} still possess a relatively high FLOPS value of 36.3G. This FLOPS requirement exceeds the capabilities of our model and does not meet the hardware requirements that we have in place.

On the other hand, our model has very low FLOPS, and its FPS and IOU scores exceed the requirements for autonomous driving scenarios. This makes our model suitable for deployment in an autonomous driving system extraction module on DRIVE PX 2 that handles 12 cameras simultaneously, where computational efficiency is crucial.

\subsection{Ablation Study }
\begin{figure}[t]
\vspace{-0mm}
  \centering
  \includegraphics[width=0.90\linewidth]{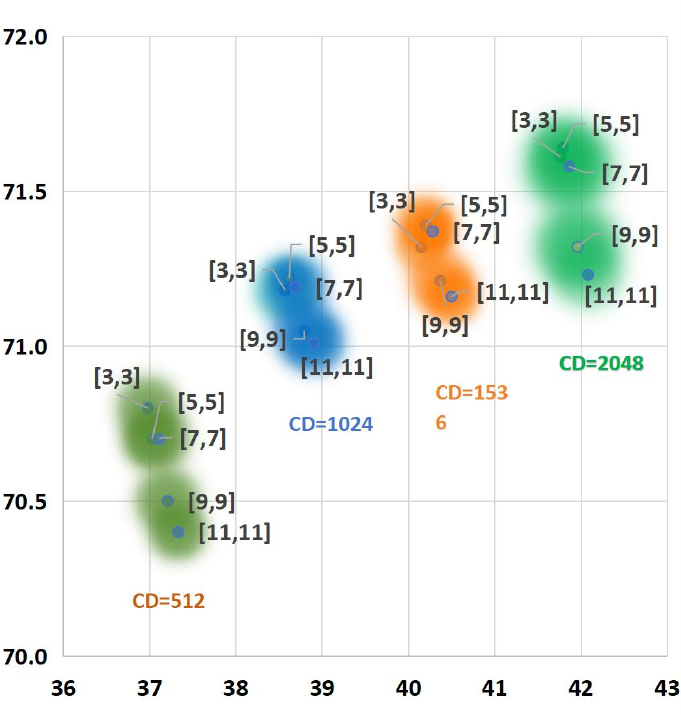}
  \caption{
Ablation results about different classifier depth and classifier kernels on CITYSCAPES. The vertical axis is mIoU and the horizontal axis is gFLOPS.}
  \label{fig:ablation}
\vspace{-5.5mm}
\end{figure}

\noindent{\textbf{Hyper-parameters in Feature Extractor.}} 
Under the condition of WIDTH\_MULTIPLIER=1, Table~\ref{tbl:city2} presents the performance evaluation data involving various configurations. Each row represents a specific set of conditions, including different CLASSIFIER DEPTH, CLASSIFIER KERNEL, and their corresponding MACs after model training, along with the mIOU value. The CLASSIFIER DEPTH has four options: 512, 1024, 1536, and 2048. For each of these options, one of the following CLASSIFIER KERNEL sizes can be chosen: [3,3], [5,5], [7,7], [9,9], or [11,11]. From Table~\ref{tbl:city}, the results of Figure~\ref{fig:ablation} can be obtained, which shows the logarithmic relationship between accuracy and Multiply-Accumulates. Generally, increasing CLASSIFIER DEPTH can increase the computational complexity of the model while potentially enhancing accuracy (mIOU).  Various CLASSIFIER KERNEL configurations lead to fluctuations in both MACs and mIOU values. For example, using [5,5] and [7,7] kernel size seems to yield higher mIOU values in most cases, although with slightly higher MACs overhead. On the other hand, [9,9] and [11,11] KERNELs slightly reduce mIOU while having higher MACs overhead, while these KERNELs are more suitable for tasks demanding wider contextual understanding, especially in situations with sizable objects or global structures.

\begin{table}[t]
\centering
\caption{Ablation Results About Different Classifier Depth and Classifier Kernel Strategies on Cityscapes for DM = 1.0.}
\resizebox{\linewidth}{!}{\begin{tabular}{cccc|ccccc|c|cc}
\toprule
\multicolumn{4}{c|}{\textbf{Classifier Depth}} & \multicolumn{5}{c|}{\textbf{Classifier Kernel}} & \multirow{2}{*}{\textbf{MACs}} & \multirow{2}{*}{\textbf{mIOU}} \\
\cmidrule(lr){1-4} \cmidrule(lr){5-9}
 512 & 1024 & 1536 & 2048 & [3,3] & [5,5] & [7,7] & [9,9] & [11,11] &  &  \\
\midrule
 $\checkmark$ &  &  &  & $\checkmark$ &  &  &  &  & \textbf{3698} & 77.8 \\
      $\checkmark$ &  &  &  &  & $\checkmark$ &  &  &  & 3703 & 77.7 \\
      $\checkmark$ &  &  &  &  &  & $\checkmark$ &  &  & 3711 & 77.7 \\
      $\checkmark$ &  &  &  &  &  &  & $\checkmark$ &  & 3721 & 77.5 \\
      $\checkmark$ &  &  &  &  &  &  &  & $\checkmark$ & 3733 & 77.4 \\
       & $\checkmark$ &  &  & $\checkmark$ &  &  &  &  & 3857 & 78.18 \\
       & $\checkmark$ &  &  &  & $\checkmark$ &  &  &  & 3862 & 78.22 \\
       & $\checkmark$ &  &  &  &  & $\checkmark$ &  &  & 3869 & 78.06 \\
       & $\checkmark$ &  &  &  &  &  & $\checkmark$ &  & 3879 & 78.19 \\
       & $\checkmark$ &  &  &  &  &  &  & $\checkmark$ & 3891 & 77.40 \\
       &  & $\checkmark$ &  & $\checkmark$ &  &  &  &  & 4015 & 78.32 \\
       &  & $\checkmark$ &  &  & $\checkmark$ &  &  &  & 4020 & 78.39 \\
       &  & $\checkmark$ &  &  &  & $\checkmark$ &  &  & 4028 & 77.87 \\
       &  & $\checkmark$ &  &  &  &  & $\checkmark$ &  & 4037 & 77.10 \\
       &  & $\checkmark$ &  &  &  &  &  & $\checkmark$ & 4050 & 76.90 \\
       &  &  & $\checkmark$ & $\checkmark$ &  &  &  &  & 4176 & 78.71 \\
       &  &  & $\checkmark$ &  & $\checkmark$ &  &  &  & 4179 & \textbf{78.80} \\
       &  &  & $\checkmark$ &  &  & $\checkmark$ &  &  & 4186 & 78.58 \\
       &  &  & $\checkmark$ &  &  &  & $\checkmark$ &  & 4196 & 78.0 \\
       &  &  & $\checkmark$ &  &  &  &  & $\checkmark$ & 4208 & 77.80 \\
\bottomrule
\end{tabular}}
\label{tbl:city2}
\vspace{-2mm}
\end{table}

In Figure~\ref{fig:ablation}, analogous outcomes are achieved when utilizing WIDTH\_MULTIPLIER values of 0.75, 0.50, and 0.25. This approach utilizes the parameters WIDTH\_MULTIPLIER, CLASSIFIER DEPTH, and CLASSIFIER KERNEL to implement a three-tier control over the computational complexity of the model. Initially, WIDTH\_MULTIPLIER is employed to broadly adapt the model size based on the available hardware capabilities. Subsequently, CLASSIFIER DEPTH is fine-tuned to achieve size adjustments within a narrower scope. Finally, CLASSIFIER KERNEL is utilized to fine-tune the model's sizing for various scenarios.

\subsection{The Overhead of Parameter Search.}
Regarding the overhead of parameter search and tuning, our tests show that the process takes only about 6 seconds. While there is a small overhead associated with the parameter search, it is designed to yield the optimal combination that maximally satisfies the computational budget while ensuring the highest possible accuracy. Although this incurs some computational cost, the significant improvement in model performance within the given constraints justifies it, making it a valuable trade-off for specific tasks or scenarios.
\color{black} 

\section{Discussion}
\color{black}Our method is designed with flexibility and efficiency in mind, which targets resource-constrained environments like embedded devices. However, scaling the model while increasing the parameter count introduces some challenges:

\begin{itemize}[leftmargin=*]
    \item \underline{Memory Constraints:} 
    As the model scales, it requires more memory for parameters and intermediate activations, which can restrict model size on memory-constrained platforms like DRIVE PX 2. Techniques such as memory-efficient architectures and parameter sharing may alleviate some burden, but trade-offs with performance must be considered.
    
    \item \underline{Energy Efficiency:} 
    Increasing model size may reduce energy efficiency. While our method aim for energy-efficient computation, scaling can lead to diminishing returns in energy-to-accuracy trade-offs. Future research should focus on incorporating low-precision computing, model pruning, and dynamic voltage scaling.

    \item \underline{Scalability Optimization:} 
    Expanding the number of parameters increases the computational demands of model optimization, potentially slowing training and inference when adjusting width multipliers, kernel sizes, and classifier depths for hardware-specific customization. Techniques like NAS or hardware-aware pruning may help balance performance with computational resources.
\end{itemize}
\noindent

To address scalability challenges, we will investigate advanced techniques such as model pruning~\cite{ji2025computation,li2025mutual,liu2025toward}, which enable neural networks to maintain computational efficiency as the number of parameters increases, particularly in resource-limited settings~\cite{liu2025rora,liu2025rcr,liu2025structured,niu2025mobile,tan2025zeroqat,zhang2025towards,tan2025harmony,10.1145/3747842,tan2025perturbation,yang2025fairsmoe,yuan2021work,yuan2022mobile,liu2023scalable}.

Furthermore, we will employ diffusion-based generative models~\cite{meng2024instructgie} to synthesize compact, informative samples~\cite{liu2023interpretable,liu2021explainable,liu2022efficient}, thereby facilitating scalable experimentation in healthcare-related tasks~\cite{liu2024brain,wang2025fairgnn,chinta2023optimization,wang2025graph,wang2025amcr,chinta2025ai}. These synthetic samples will be integrated with large language models (LLMs) to support downstream applications such as 3D reconstruction~\cite{lei2023mac,li2025local,dong2024physical,dong2024df}, enhancing overall model utility in multi-modal scenarios.

%% file: sections/7-conclusion.tex
\section{Conclusion}

In conclusion, our approach effectively addresses the challenges faced by autonomous driving platforms by customizing semantic segmentation networks to accommodate varying hardware resources and precision requirements. By implementing a three-tier control mechanism—width multiplier, classifier depth, and classifier kernel—we achieve dynamic adaptability that facilitates broad model scaling and targeted refinement. Using Bayesian Optimization with surrogate modeling allows efficient exploration of hyperparameter spaces within tight computational budgets, ensuring optimal performance and resource allocation for autonomous driving tasks.

The ability to scale Multiply-Accumulate Operations (MACs) for Task-Specific Learning Adaptation (TSLA) maximizes computational capacity and model accuracy, enhancing hardware utilization. Looking ahead, we plan to apply the TSLA method to the efficient design of medical imaging and real-time human body tracking systems, further demonstrating the versatility of our approach in critical applications.reconstruction~\cite{dong2024physical,dong2024df} models.